%% file: main.tex
\documentclass[letterpaper, 10 pt, conference]{ieeeconf}  

\IEEEoverridecommandlockouts                              
\usepackage[dvipsnames]{xcolor}
\usepackage[pdftex]{graphicx} 
\usepackage{cite}
\usepackage{graphics} 
\usepackage{epsfig} 
\usepackage{times} 
\usepackage{amsmath} 
\usepackage{amssymb}  
\usepackage{subcaption}
\usepackage{siunitx} 
\usepackage{booktabs} 
\usepackage{makecell}
\usepackage{multirow}
\usepackage{color, colortbl}
\usepackage{adjustbox}
\usepackage{esvect}
\usepackage{tikz}
\usepackage{float} 
\usepackage{algpseudocode}
\usepackage{algorithm}
\usepackage{tabularx}
\usepackage{svg}
\usepackage{adjustbox}

\usepackage{accents}  
\usepackage{url}
\usepackage{threeparttable} 

\def \measurements{Z}

\definecolor{Gray}{gray}{0.9}

\graphicspath{{fig/}}
\title{\LARGE \bf Semantic Enhancement for Object SLAM with Heterogeneous Multimodal Large Language Model Agents}
\author{Jungseok Hong$^{1*}$, Ran Choi$^{1*}$, John J. Leonard$^{1}$ 
\thanks{$^{1}$ The authors are with the Computer Science and Artificial Intelligence Laboratory (CSAIL) at the Massachusetts Institute of Technology (MIT), 32 Vassar St, Cambridge, MA 02139, USA.
        Corresponding author: Jungseok Hong ({\tt\small jungseok@mit.edu}) * indicates equal contribution.}%
        }

\begin{document}

\maketitle
\thispagestyle{empty}
\pagestyle{empty}
\input{src/0-abstract}
\input{src/1-intro}
\input{src/2-relwork}

\input{src/4-method}
\input{src/5-experiments}
\input{src/7-conclusion}

\bibliographystyle{IEEEtran}
\bibliography{reference.bib}

\end{document}

%% file: src/0-abstract.tex
\begin{abstract}
Object Simultaneous Localization and Mapping (SLAM) systems struggle to correctly associate semantically similar objects in close proximity, especially in cluttered indoor environments and when scenes change. 
We present Semantic Enhancement for Object SLAM (SEO-SLAM), a novel framework that enhances semantic mapping by integrating heterogeneous multimodal large language model (MLLM) agents. 
Our method enables scene adaptation while maintaining a semantically rich map. To improve computational efficiency, we propose an asynchronous processing scheme that significantly reduces the agents' inference time without compromising semantic accuracy or SLAM performance.
Additionally, we introduce a multi-data association strategy using a cost matrix that combines semantic and Mahalanobis distances, formulating the problem as a Linear Assignment Problem (LAP) to alleviate perceptual aliasing.
Experimental results demonstrate that SEO-SLAM consistently achieves higher semantic accuracy and reduces false positives compared to baselines, while our asynchronous MLLM agents significantly improve processing efficiency over synchronous setups. We also demonstrate that SEO-SLAM has the potential to improve downstream tasks such as robotic assistance.
Our dataset is publicly available at: \url{jungseokhong.com/SEO-SLAM}.
\end{abstract}

%% file: src/1-intro.tex
\section{Introduction}\label{sec:intro}
Semantic understanding is crucial for robots operating in real-world environments. Traditional Simultaneous Localization and Mapping (SLAM)~\cite{cadena2016past} systems primarily focus on geometric accuracy, but without semantic awareness, they struggle with essential tasks such as distinguishing objects, reasoning about their relationships, or adapting to scene changes. 
For instance, autonomous robots navigating cluttered spaces must differentiate between visually similar objects, such as different types of shoes~(Fig~\ref{fig:intro}) and books. The robots also need to update their internal maps when scenes change. The ability to incorporate rich, dynamic semantic information into SLAM pipelines is key to enabling more intelligent, adaptable robotic perception.

\input{figs/intro_2025}

We present SEO-SLAM (Semantic Enhancement for Object SLAM), a novel framework that integrates heterogeneous multimodal large language model (MLLM) agents to enhance object-based semantic SLAM. Instead of relying on a single large MLLM model to handle all semantic reasoning tasks, our approach leverages multiple specialized MLLM agents, each assigned to a specific function, such as object verification, label refinement, and landmark merging. These agents operate asynchronously, meaning they do not have to wait for one another to complete inference before proceeding. This design significantly improves computational efficiency compared to approaches that depend on a single large MLLM model handling all tasks sequentially. By distributing tasks among smaller, specialized agents, our system not only runs faster but also allows each agent to focus on a specific task, improving semantic mapping accuracy. 

\input{figs/diagram_fig}

To further enhance the efficiency of our framework, we introduce an asynchronous processing scheme that reduces the overall agent runtime without impeding the main SLAM pipeline or sacrificing semantic accuracy. Additionally, we adopt a multi-data association strategy that mitigates perceptual aliasing by combining semantic similarity scores and Mahalanobis distances into a cost matrix, which is then solved as a Linear Assignment Problem (LAP). 

While recent advancements in computer vision~\cite{cadena2016past}, foundation models~\cite{touvron2023llama, brown2020language, caron2021emerging, radford2021learning, liu2024visual, openai2024gpt4technicalreport}, and open-vocabulary object detection have significantly improved SLAM’s ability to incorporate semantic information~\cite{salzmann2024scene, chen2024spatialvlm, fu2024scene, deng2024opengraph, kerr2023lerf, matsuzaki2024clip, gu2024conceptgraphs, kassab2024language}, existing approaches still face three key challenges: (1) Perceptual aliasing, where similar objects are mislabeled or confused due to noisy measurements, (2) Outdated landmarks, as traditional SLAM systems struggle to update maps when objects move or disappear, and (3) Computational bottlenecks, since existing methods typically rely on a single model for all semantic reasoning, leading to high latency and inefficient processing. SEO-SLAM overcomes these challenges by enabling multiple heterogeneous MLLM agents to work asynchronously, refining object labels, and removing erroneous landmarks rather than directly use MLLM models to detect objects.

Our main contributions are as follows:
\begin{itemize}
    \item A novel object-based semantic SLAM framework that integrates heterogeneous MLLM agents to maintain a semantically rich and dynamically updated map.
    \item An asynchronous MLLM-agent processing scheme that significantly improves computational efficiency while maintaining high accuracy.
    \item A multi-data association method that combines semantic similarity scores and Mahalanobis distances, formulating the problem as a Linear Assignment Problem (LAP) for reducing perceptual aliasing.
    \item Experimental evaluations demonstrating our method's capabilities of semantically rich object mapping and landmark updating in challenging scenarios.
\end{itemize}

%% file: figs/intro_2025.tex
\begin{figure}[t!]
\vspace{2mm}
        \centering
        \includegraphics[width=1\linewidth]{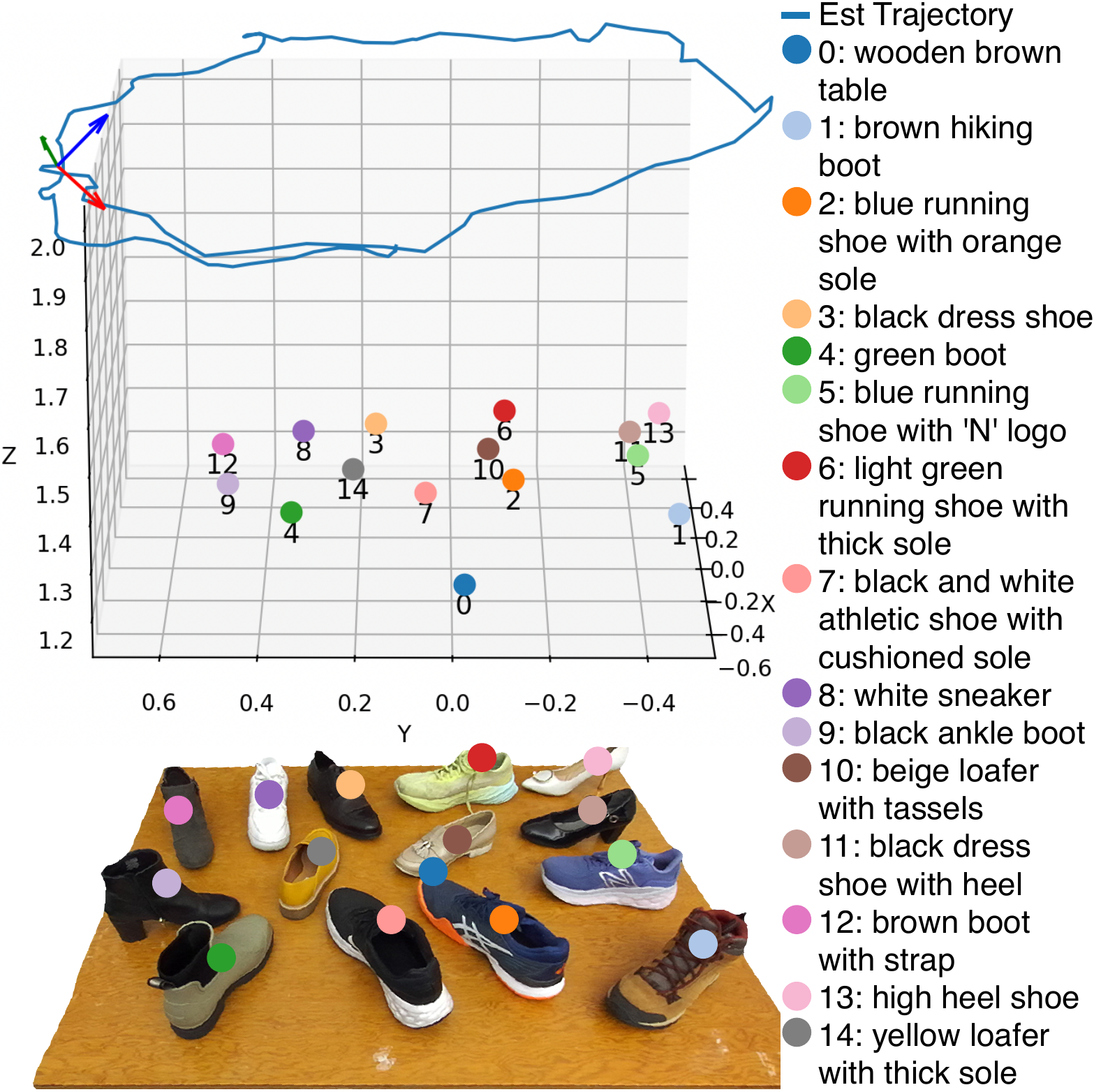}
        \caption{A sample semantic mapping result by SEO-SLAM. The plot shows a 3D map of the scene in the RGB image, including landmark positions and a camera trajectory, where numbers/colors represent class labels shown in the legend. Each shoe is assigned descriptive labels capturing their visual characteristics created by our proposed MLLM agents.}
        \vspace{-4mm}
        \label{fig:intro}
\end{figure}

%% file: figs/diagram_fig.tex
\begin{figure*}  
    \vspace{2mm}
    \centering
    \includegraphics[width=1.0\linewidth]{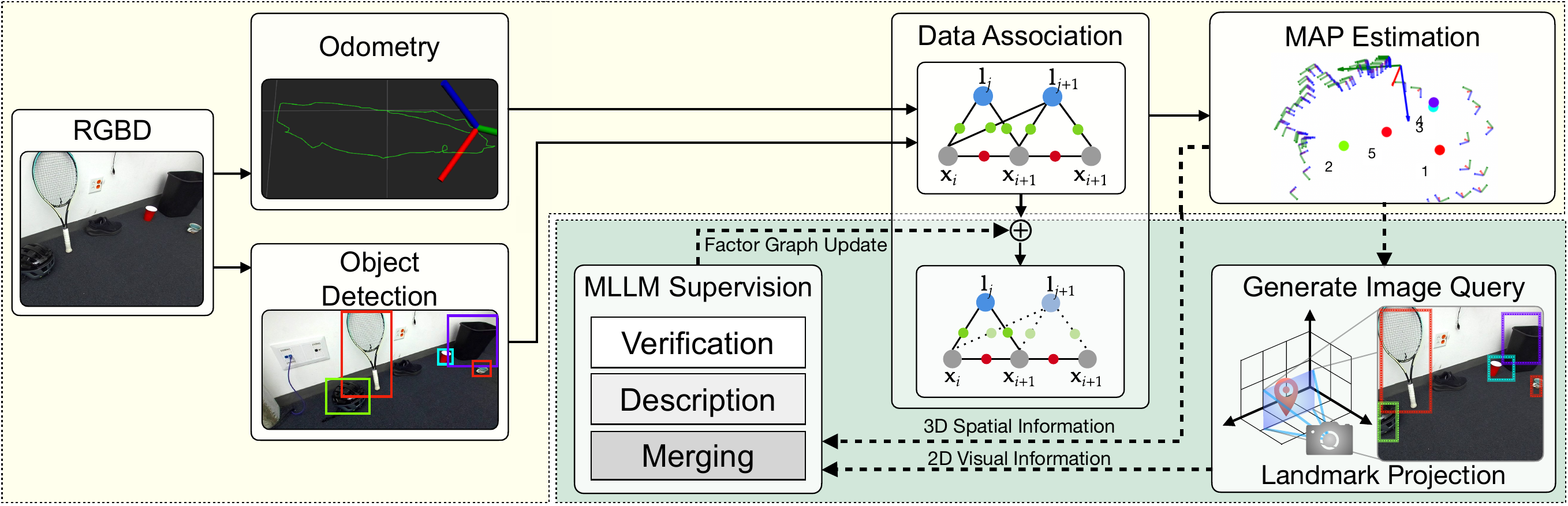}
    \caption{Overview of the SEO-SLAM system: The pipeline begins with an RGBD image input, from which odometry is derived. Object measurements from an open-vocabulary detector are fed into a factor graph. MAP estimation is obtained through factor graph optimization. Landmarks from the current map are projected onto the camera frame and overlaid on the image. This composite image is used as input for the MLLM agents, which evaluate each landmark based on the current scene. For example, if the agents informs $\textbf{l}_{j+1}$ is an erroneous landmark, factors related to $\textbf{l}_{j+1}$ are removed. The yellow-colored region represents the primary SLAM pipeline, while the green-colored region denotes an asynchronous process that does not impact MAP runtime.} 
    \label{fig:pipeline}
    \vspace{-4mm}
\end{figure*}


%% file: src/2-relwork.tex
\section{Related Work}
Semantic SLAM extends traditional SLAM by incorporating semantic information into the mapping and localization process.
Since the early work of Galindo~\textit{et al.}~\cite{galindo2005multi}, semantic SLAM has benefited from advances in deep learning and computer vision~\cite{chen2022semantic}. Recent systems like Kimera~\cite{rosinol2020kimera}, EAO-SLAM~\cite{wu2020eao}, and DSP-SLAM~\cite{wang2021dsp} have demonstrated improved performance in various environments.
However, these systems often struggle with scene changes and distinguishing between semantically similar objects.

Some approaches, like~\cite{yu2018ds} and~\cite{xiao2019dynamic}, have attempted to address the challenge of dynamic objects by incorporating prior knowledge about those objects.
Several object-centric SLAM approaches, such as~\cite{iqbal2018localization, wu2020eao,adkins2024obvi}, have focused on improving object-level mapping accuracy. 
These methods primarily evaluate their performance based on the quantity of objects correctly mapped. However, they are generally limited to closed-set object scenes and lack evaluation of semantic accuracy, particularly in distinguishing between similar objects or handling precise classification.

Data association is crucial for handling perceptual aliasing and maintaining map consistency in SLAM. Early works addressed this with approaches such as joint compatibility~\cite{neira2001data}, probabilistic data association~\cite{BARSHALOM1975451}, multiple hypothesis~\cite{reid1979algorithm}.
As semantic SLAM has emerged, data association techniques that use semantic information have been introduced. Bowman~\textit{et al.}~\cite{bowmanslam} proposed an Expectation-Maximization (EM) algorithm-based approach to jointly optimize landmark position and semantics. Utilizing mixture models~\cite{olson2013inference}, Doherty \textit{et al.}~\cite{kevin19slam} coupled geometric (continuous) and semantic (discrete) information of landmarks for optimization, where they show that a mixture approach outperforms the EM-based approach. Bernreiter~\textit{et al.}~\cite{bernreiter2019multiple} developed a multiple hypothesis approach with an optimizing hypothesis tree. 
However, these methods are often limited to closed-set semantics or require prior knowledge of a multiclass prediction confusion matrix.

The recent advances in foundation models~\cite{openai2024gpt4technicalreport} and open-vocabulary detection~\cite{zareian2021open} have advanced semantic SLAM such as ConceptGraphs~\cite{gu2024conceptgraphs} and LEXIS~\cite{kassab2024language}. 
Gu~\textit{et al.}~\cite{gu2024conceptgraphs} focus on building a semantic map embedding multimodal features for various downstream tasks. Kassab~\textit{et al.}~\cite{kassab2024language} show how open-vocabulary language models can be integrated into SLAM systems to enhance semantic understanding. However, the challenges in semantic SLAM introduced in Section~\ref{sec:intro} still remain an open problem.

SEO-SLAM builds on the factor graph framework GTSAM~\cite{kaess2012isam2}, incorporating heterogeneous MLLM models to generate semantically rich landmark labels, construct a consistent semantic map, and eliminate erroneous landmarks.
With this approach, we aim to address the limitations of existing semantic SLAM systems, particularly in environments with multiple similar objects, and enable more accurate and consistent mapping. 

%% file: src/4-method.tex
\section{Method}
Object SLAM can be defined as finding a set of robot poses $\textbf{x}_i$ and landmark positions $\textbf{l}_j$ given a set of measurements $\textbf{z}_k$. 
This can be formulated as a maximum a posteriori (MAP) problem, and $\hat{\textbf{x}}$ represents an estimated state vector including robot pose $\hat{\textbf{x}}_i$ and landmark $\hat{\textbf{l}}_j$.
\begin{equation}
\hat{\textbf{x}}=\underset{\textbf{x}}{\arg \max } \; p(\textbf{x} \mid \measurements)
\end{equation}
where $\textbf{x}_i \in SE(3), \textbf{l}_j \in\mathbb{R}^{3}$, and $Z = \{\textbf{z}_k\}$.


We associate both semantic and geometric measurements to landmarks and optimize the MAP problem.
Our approach supports open-vocabulary semantic classes without needing prior knowledge of class prediction statistics from a multiclass prediction confusion matrix. 
Fig~\ref{fig:pipeline} shows our SEO-SLAM pipeline. It takes an RGBD image as an input and obtains odometry from visual odometry. An RGB image is fed into an open vocabulary detector. From detection results and a depth image, we extract geometric and semantic measurements of detected objects. They are used for multi-data association to estimate landmarks and robot poses by optimizing a factor graph. For the MLLM agents to evaluate the landmarks, we provide their 3D coordinates and an image, onto which each landmark is projected to a current camera frame.
The output from the agents is then used to refine the map by updating a factor graph. Each component of the pipeline is detailed in the following sections.

\subsection{Object Measurements}\label{sec:om}
We employ an object detection pipeline, which can detect objects in an open-vocabulary setting. We use RAM-GroundingDINO (RG)~\cite{ren2024grounded} for object detection, which combines two models: (1) Recognize-Anything Model (RAM)~\cite{huang2023inject}, an open-set tagging model, and (2) GroundingDINO~\cite{liu2023grounding}, a grounding model for object detection. 
The initial detection labels are used by our MLLM agents to improve each object's description. 
From each detection, we store:
\begin{itemize}
    \item Object centroid $(c_x,c_y)$
    \item Bounding box width $o_W$ and height $o_H$
    \item Class label $cl$
    \item Feature embedding $\phi(cl)$ using word embedding
    \item Object centroid position in 3D $(x,y,z)$ derived from camera intrinsics and depth image.
\end{itemize}
From these, we form a measurement model of the landmark's position relative to the robot:
\begin{equation}\label{eq:model}
    \textbf{z}_{ij} := h_{ij}(\textbf{x}) + \textbf{v}_{ij}
\end{equation}
where $\textbf{v}_{ij}$ is a zero-mean normal distribution noise with covariance $\Gamma$.
$h_{ij}(\textbf{x})$ projects landmark $\textbf{l}_j$ into the coordinate frame of robot pose $\textbf{x}_i$. $\Gamma$ encapsulates both sensor uncertainty and any approximation error in the object detection stage.
The measurement model related probability is
\begin{equation}\label{eq:ml}
    P({\textbf{z}}_k, j_k = j | \textbf{x}) = \frac{1}{\sqrt{|2\pi \Gamma|}} e^{-\frac{1}{2} \|\mathbf{h}_{ikj}(\textbf{x}) - {\textbf{z}}_k\|^2_{\Gamma}}
\end{equation}

\subsection{Multi-Data Association} 
Given the model and a new measurement $\textbf{z}_k$, our aim is to determine the probability $P(\textbf{z}_k, j_k = j | Z^-)$, which captures the likelihood of $\textbf{z}_k$ to landmark $j$ associations, considering all prior measurements $Z^-$. 
As derived in~\cite{kaess2009covariance}, this probability can be connected to SLAM with a state vector $\textbf{x}$.
When $\textbf{x}$ has an associated covariance $\Sigma$, we linearize $h_{ij}$ around the current estimate $\hat{\textbf{x}}$. This yields a total measurement covariance
\begin{equation}\label{eq:cov}
    C_{i_{k}j} := \frac{\partial \mathbf{h}_{i_{k}j}}{\partial \textbf{x}} \Big|_{\hat{\textbf{x}}} \Sigma \frac{\partial \mathbf{h}_{i_{k}j}}{\partial \textbf{x}} \Big|_{\hat{\textbf{x}}}^T + \Gamma
\end{equation}
Additionally, the state prior around the mean $\hat{\textbf{x}}$ can be described by
\begin{equation}\label{eq:cse}
    P(\textbf{x} | Z^-) = \frac{1}{\sqrt{|2\pi \Sigma|}} e^{-\frac{1}{2} \|\textbf{x} - \hat{\textbf{x}}\|^2_{\Sigma}}
\end{equation}
Given~Eq~\ref{eq:ml}-\ref{eq:cse}, the measurement likelihood can be approximated by
\begin{equation}\label{eq:approx}
\begin{split}
        P({\textbf{z}}_k, j_k = j | Z^-) &= \int_\textbf{x} P({\textbf{z}}_k, j_k = j | \textbf{x}) P(\textbf{x} | Z^-)d\textbf{x}\\
    &\approx \frac{1}{\sqrt{|2\pi C_{ikj}|}} e^{-\frac{1}{2} \|\mathbf{h}_{i_{k}j}(\hat{\textbf{x}}) - {\textbf{z}}_k \|^2_{C_{i_{k}j}}}
\end{split}    
\end{equation}

\input{figs/agent_workflow}

\input{figs/prompt_example}

Taking a negative log on Eq~\ref{eq:approx} yields a chi-square test~\cite{kaess2009covariance} in Eq~\ref{eq:chi1}. 
\begin{equation}\label{eq:chi1}
    D_{kj}^{2, ML} := \|\mathbf{h}_{ikj}(\hat{x}) - {\textbf{z}}_k\|_{C_{ikj}}^2 < \chi_{d, \alpha}^2
\end{equation}
where $\chi_{d, \alpha}^2$ is a chi-square threshold for $d$ degrees of freedom and confidence $\alpha$.
We use this to determine if there is a significant geometric association between a given landmark and measurements. 
To make the data association scalable, we only run the test with the landmarks within the search distance $d$ from robot pose $\hat{\textbf{x}}$.
Once measurement $\textbf{z}_k$ passes the chi-square test for landmark $\hat{\textbf{l}}_j$, we also compute a semantic similarity score using the feature embedding
\begin{equation}
    {s} = \phi(\textbf{z}_k) \cdot \phi(\hat{\textbf{l}}_j)
\end{equation}
If $s$ is above threshold $\tau_s$ then we assign the Mahalanobis distance $D_{kj}^{2, ML}$, as a cost to ($\textbf{z}_k$, $\textbf{l}_j$) pair. Otherwise, we set a predefined large cost for the pair. 
In our preliminary tests, we observed that having two stages (\textit{e.g.,} chi-square and semantic similarity tests) is more robust to perceptual aliasing than combining them together in a one weighted function.
In this way, we calculate the costs for $K\times N$ pairs when there are $K$ measurements and $N$ landmarks. To solve it as a linear assignment problem (LAP), we build a $K \times (N+K)$ cost matrix, which is initialized with the predefined maximum values. 
Upon evaluating the cost, we update the cost to the first N columns. The remaining $K$ columns are used to allow measurement $\textbf{z}_k$ to be assigned to ``out-of-range landmark", which indicates there is no matching landmark. 
This ensures consistent, one‐to‐one matching, out‐of‐range landmark assignments, and rejection of false positive measurements.
We adopt the Jonker-Volgenant-Castanon (JVC)~\cite{jonker1988shortest} algorithm to assign each measurement to landmarks by solving the LAP. When the assigned index, $idx$, is smaller or equal to the number of landmarks, $\textbf{z}_k$ is associated with $\textbf{l}_{idx}$. Otherwise, a new landmark is created for that measurement.

\subsection{Landmark Projection}\label{sec:lp}
Once the main SLAM pipeline estimates a map, it is essential to extract information in a format that enables MLLM agents to reason about landmarks effectively. A straightforward approach is to provide the 3D coordinates of landmarks alongside 2D images from the current camera view. 
However, this approach often struggles with hallucinations and may fail to capture contextual relationships that arise from integrating 2D and 3D information.
To address this, we propose landmark projection, which overlays the map’s 3D data onto the camera’s 2D image, forming a composite view. This method is motivated by~\cite {yang2023set, yang2023dawn}, which demonstrates that visually marking object locations on images enhances an MLLM’s spatial understanding. 

Specifically, once robot pose $\hat{\textbf{x}}_i$ and landmark positions $\hat{\textbf{l}}$ are estimated, we first transform each landmark position to the camera frame $(x_C, y_C, z_C)$ and project the positions on the image plane and estimate $(u_C,v_C)$.
If $(u_C,v_C)$ lies within the image plane, then we compare $z_C$ with $z_D$, which is the depth of a corresponding depth image at $(u_C,v_C)$. When $z_C <z_D$, we use the bounding box size ($o_W \times o_H$ ) and depth $Z$ obtained from object measurements (Section~\ref{sec:om}) to generate an image where the estimated landmark bounding boxes are overlaid on the RGB image~(Fig~\ref{fig:api_input}). 
The MLLM agents use this projected information and 3D coordinates of landmarks as an input to evaluate each landmark.

\input{figs/system_prompt}

\subsection{MLLM agents}
We introduce a team of heterogeneous MLLM agents that improve a semantic map from a main SLAM pipeline without compromising its runtime. 
Given the input described in Section~\ref{sec:lp}, each agent outputs landmark ``key'' to refine and eliminate factors from the factor graph.  
Our approach leverages three specialized MLLM agents (Fig~\ref{fig:api_diagram}): (1) \textit{Verification Agent}, which assesses landmark labels and identifies false positives; (2) \textit{Description Agent}, which generates semantically rich labels from visual features of given landmarks to enhance overall semantic information; and (3) \textit{Merging Agent} which detects duplicate landmarks of the same object and merges them using both global and local context.
These agents operate asynchronously with a fixed number of threads, enabling parallel processing of object queries. 
This design improves both processing speed and response time compared to synchronous approaches while ensuring efficient landmark management under the token-per-minute (TPM) and request-per-minute (RPM) constraints of the MLLM models.

\subsubsection{Verification Agent} assesses the correspondence between existing landmark labels and visual landmark information by analyzing cropped projection images with textual queries (Fig~\ref{fig:api_input}a). This agent evaluates each label’s accuracy, categorizing it as \textit{correct, incorrect}, or \textit{unidentifiable}~(Fig~\ref{fig:api_diagram}).
To efficiently manage the increasing number of landmarks over time, we design this verification process as a classification task, allowing the use of smaller models compared to the other agents handling more complex operations. For enhanced efficiency, each query prompt for a thread includes up to four projected landmarks, enabling simultaneous evaluation of multiple landmarks.

\subsubsection{Description Agent} recognizes objects, describes their characteristics, and generates new labels for both newly added landmarks and existing ones with incorrect labels (Fig~\ref{fig:api_diagram}). Labels need to capture distinctive object features with consistency—identical objects should receive identical labels. Therefore, they are generated based on fundamental attributes such as color, shape, and special characteristics to ensure semantic richness.
This agent processes one landmark projection per query, allowing focused analysis of individual object characteristics. 
Additionally, the agent maintains a list of previously generated labels~(Fig~\ref{fig:api_input}b) and compares newly generated labels with the ones in the list. It assigns matching labels from the list when semantic equivalence is found; otherwise, it allocates new labels~(Fig~\ref{fig:api_prompt}) to increase label consistency.

\subsubsection{Merging agent} handles duplicate landmarks that may have different labels for the same object. Due to false positives in object detection and odometry noises, multiple landmarks may exist for a single object under various labels, such as ``office supply and scissors" or ``toy and figure."
This agent processes two categories of landmarks: (1) those validated by the Verification Agent with confirmed correct labels, and (2) those newly labeled by the Description Agent~(Fig~\ref{fig:api_diagram}). Since this task requires both global and local contexts, the agent receives an entire scene with landmarks, cropped images of individual landmarks, and textual queries containing 3D coordinates of the landmarks~(Fig~\ref{fig:api_input}c).
Using this comprehensive information, the agent identifies and merges redundant landmarks based on their semantic relationships, spatial information, and visual appearances~(Fig~\ref{fig:api_prompt}).
\input{figs/dataset_tab}

\subsection{System Prompt Design for MLLM Agents}
Each MLLM agent operates with carefully structured instruction prompts that define its role, tasks, and standardized response format~(Fig~\ref{fig:api_prompt}), where each instruction is segmented into step-by-step guidelines to enhance reliability~\cite{wei2022chain, zhang2022automatic}.
This approach maintains the consistency of the outputs in different scenes. 
We focus on obtaining reliable responses from the agents under this structured prompting strategy and avoid fine-tuning them to make our approach more generalizable.

%% file: figs/agent_workflow.tex
\begin{figure}[t!]
\vspace{2mm}
    \centering
    \includegraphics[width=0.9\linewidth]{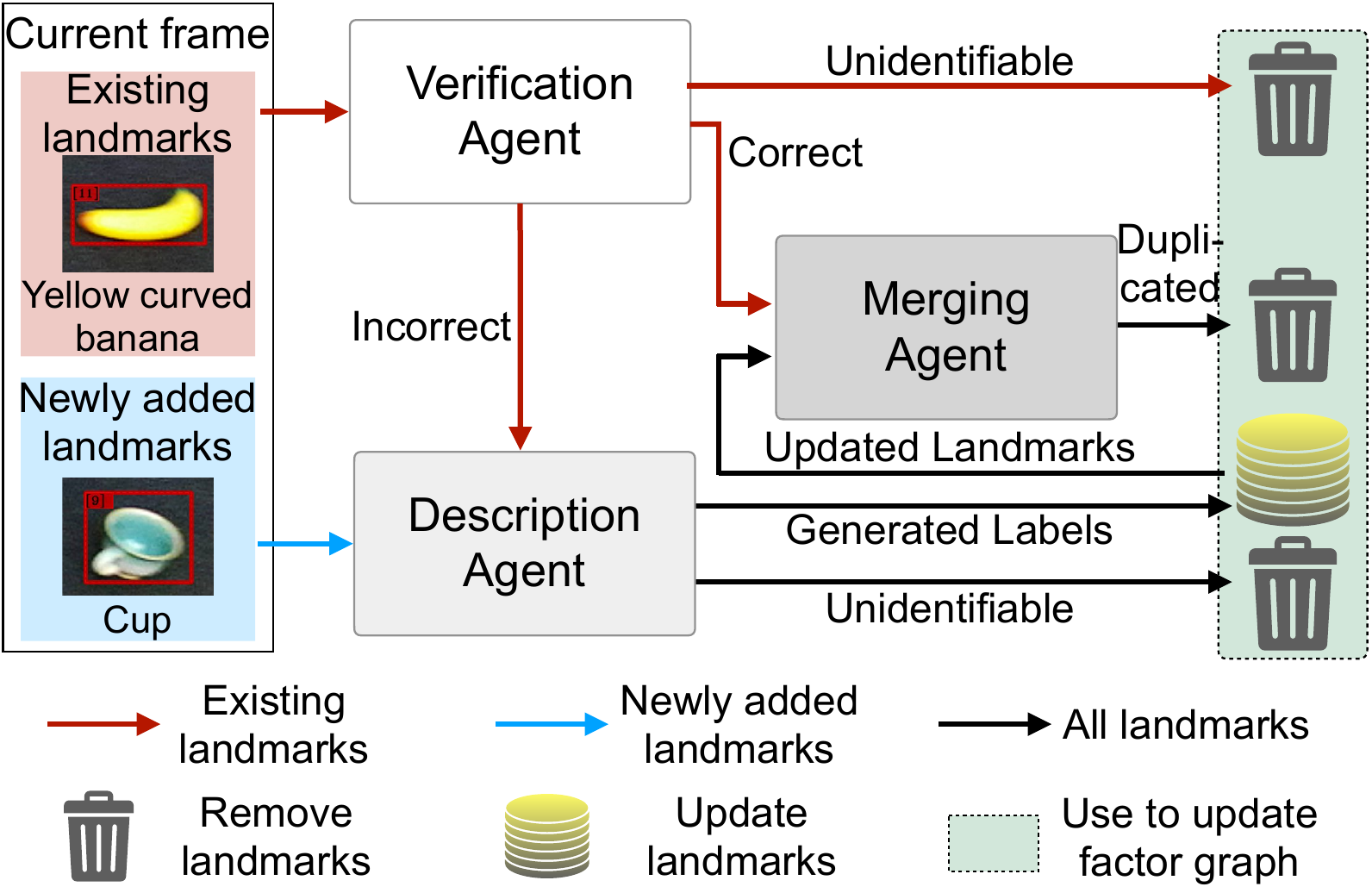}
    
    \caption{Workflow of the MLLM agents in SEO-SLAM. Landmarks are introduced from the current frame (existing or newly added) and pass through the Verification Agent, which flags incorrect or unidentifiable labels. The Description Agent assigns or refines semantic labels for new and existing landmarks while removing unidentifiable entries. Finally, the Merging Agent merges duplicates.}
    \vspace{-4mm}
    \label{fig:api_diagram}
\end{figure}

%% file: figs/prompt_example.tex
\begin{figure}[t!]
\vspace{2mm}
    \centering
    \includegraphics[width=1\linewidth]{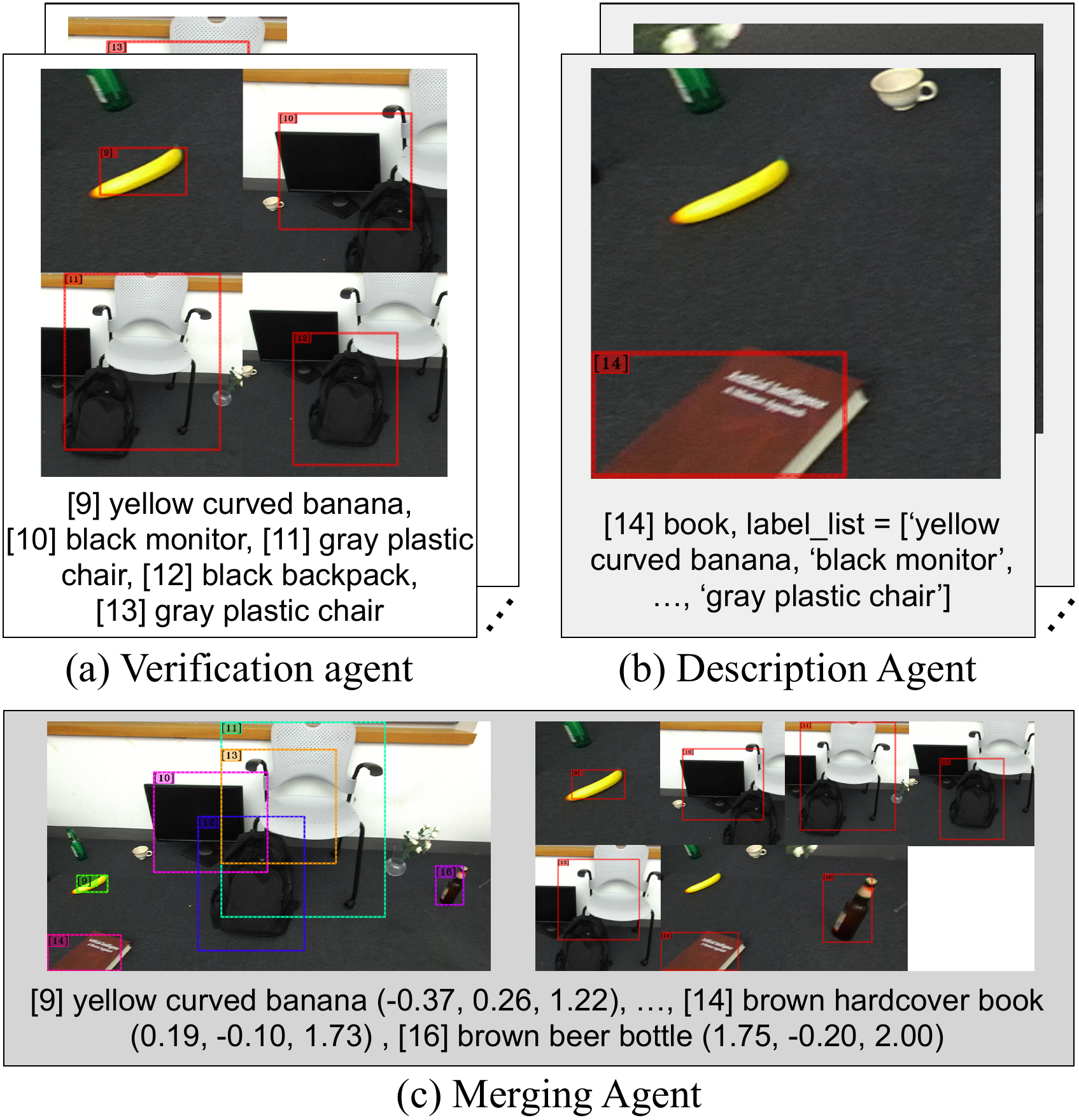}
    \caption{Query prompts examples for each agent in SEO-SLAM. Each query consists of a projection image and a list of landmarks in the format `[$<$index$>$] $<$object label$>$'. This example includes seven landmarks, with five newly detected and two existing ones. (a) The Verification Agent evaluates landmark labels by analyzing a composite image (512x512) containing up to four cropped projected landmarks per thread. (b)The Description Agent refines object labels by inspecting individual landmark projections (512×512) per thread. (c) The Merging Agent processes a full-scene projection (512×512), individual landmark projections (256×256), and each landmark’s 3D coordinates to detect and merge duplicate landmarks. Each query thread includes one full-scene image and up to eight individual landmark projections. Parentheses indicate image sizes.}
    \vspace{-4mm}
    \label{fig:api_input}
\end{figure}

%% file: figs/system_prompt.tex
\begin{figure}[btp]
\vspace{2mm}
    \centering
    \includegraphics[width=1\linewidth]{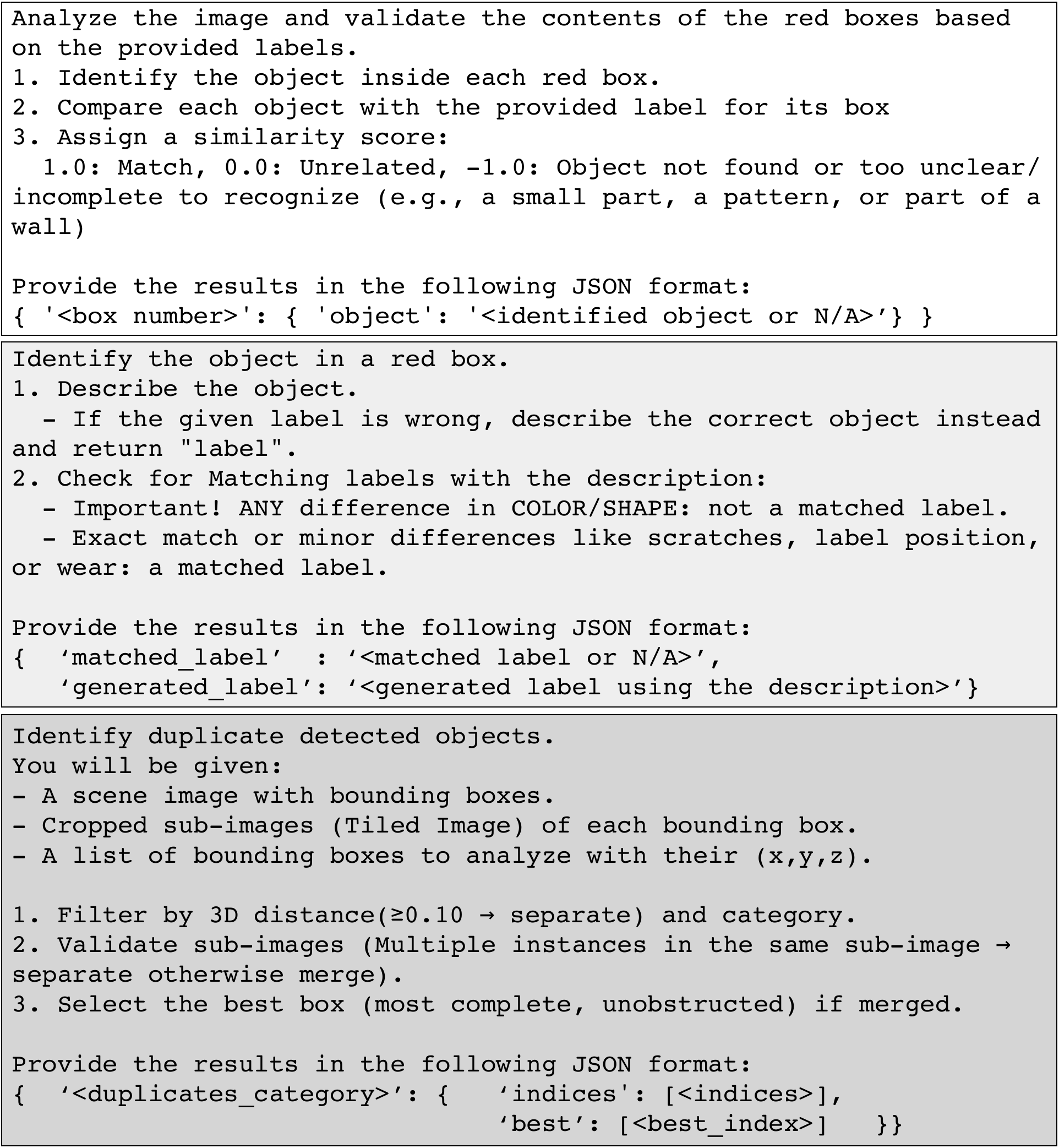}
    \caption{Simplified System prompts for: Verification Agent~(top),  Description Agent~(middle), and Merging Agent~(bottom).}
    \vspace{-6mm}
    \label{fig:api_prompt}
\end{figure}

%% file: figs/dataset_tab.tex

\begin{table}[t]
\vspace{2mm}
    \centering
        \caption{Summary of Object Datasets*}\label{tab:dataset}
    \begin{tabular}{cp{6cm}} 
        \toprule
        \textbf{Dataset} & \multicolumn{1}{c}{\textbf{Objects}} \\ \midrule
        Small 1 (11) & bag, banana, book, chair, fan, flower, helmet, scissors, shoe, racquet, vase \\ \midrule
        Small 2 (11) & banana, bin, book, chair, fan, flower, football, scissors, snowman, teacup, vase \\ \midrule
        Medium 1 (21) & bag (2), banana (2), book (2), bottle (2), chair (2), fan (2), flower, helmet, racquet, scissors, shoe (2), snowman, toy car, vase \\ \midrule
        Medium 2 (20) & banana (2), bin (2), book (2), chair (2), cup, fan, flower, football, helmet, scissors, shoe, snowman, teacup (2), racquet, vase \\ \midrule
        Large 1 (32)  & bag, banana (3), bin (3), book (3), bottle (2), chair (3), cup, fan, flower, football, helmet, monitor (3), electric outlet, racquet, scissors, shoe (2), snowman, teacup (2), vase \\ \midrule
        Large 2 (29)  & apple (2), bag, banana (3), bin (2), book, bowl, chair (2), Cheez-It box (2), electric outlet, fan, flower, football (2), hammer (2), helmet, racquet, scissors, shoes(3), teacup, vase \\ \bottomrule
    \end{tabular}
    \begin{tablenotes}\footnotesize
\item[*] (*) indicates the number of objects
\end{tablenotes}
\vspace{-4mm}
\end{table}

%% file: src/5-experiments.tex
\section{Experimental Setup and Results}

\subsection{Datasets for Evaluations}

To evaluate our proposed SEO-SLAM framework, we collected six indoor datasets with varying object densities, categorized as small ($\approx 10$ objects), medium ($\approx 20$ objects), or large ($\approx 30$ objects) (Table~\ref{tab:dataset}).
These datasets include common office and household items such as chairs, books, bottles, and shoes, with objects placed in close proximity to introduce perceptual aliasing challenges. Each dataset also includes scene changes, where objects are removed or repositioned to assess the system’s adaptability to dynamic environments.
We use a ZED2i stereo camera to capture RGB images and odometry data. Additionally, we evaluate SEO-SLAM’s tracking performance on TUM RGB-D dataset~\cite{sturm12iros} sequences (fr1desk, fr1desk2, and fr2desk) to benchmark its pose estimation accuracy. These sequences were selected due to the close proximity of objects in their scenes, which presents challenges for object SLAM.
\input{figs/results2_tab}

\input{figs/results_tum}

\input{figs/stata_qual}

\subsection{Experiments Setup}



SEO-SLAM deploys RAM++ (plus\_swin\_large) and GroundingDINO (swinb\_cogcoor) for object detection, setting both the detection confidence and GroundingDINO IoU threshold to 0.5.
For the MLLM agents, we utilize:
\begin{itemize}
    \item Verification Agent: Llama-3.2-90B-Vision~\cite{touvron2023llama} for lightweight label verification to maintain efficiency.
    \item Description Agent: GPT-4o~\cite{openai2024gpt4technicalreport} to refine and generate object labels with enhanced semantic detail.
    \item Merging Agent: GPT-4o to detect and merge duplicate landmarks.
\end{itemize}
All agents are configured with a temperature of 0.7 and a maximum token limit of 250, with all other parameters set to default values. 

To optimize performance, we implement a semaphore-based asynchronous execution strategy, limiting the number of parallel threads to manage agent queries efficiently. 
This ensures that most queries are processed while remaining within Token-per-Minute (TPM) and Request-per-Minute (RPM) constraints.

Our primary focus is not on improving tracking accuracy but rather on enhancing semantic mapping accuracy, ensuring that the SLAM system produces a more meaningful and dynamically updated representation of the environment. However, to demonstrate that SEO-SLAM maintains reliable localization performance while improving semantic understanding, we evaluate Absolute Pose Error (APE) on the TUM RGB-D dataset. This comparison helps validate that our method does not significantly degrade trajectory estimation despite introducing additional semantic processing steps. We conduct a qualitative evaluation on a $\approx 250m$-long trajectory in the MIT Stata Center to further assess the system’s performance at a larger scale.

Additionally, we conduct an ablation study to analyze the contributions of each SEO-SLAM component by testing four different configurations: (1) RG, a baseline SLAM pipeline using only the open-vocabulary detector without MLLM agents or multi-data association; (2) RG+D, incorporating multi-data association to enhance robustness against perceptual aliasing; (3) RG+A, integrating MLLM agents to improve semantic label accuracy; and (4) RG+D+A, the full SEO-SLAM pipeline, which combines multi-data association and MLLM agents for optimal performance. 
This structured evaluation enables a systematic assessment of each component’s contribution to the accuracy of the generated semantic maps.

\input{figs/results_time}

\subsection{Results}
The results demonstrate that SEO-SLAM significantly enhances semantic mapping accuracy while maintaining competitive pose estimation performance. Our ablation study (Table~\ref{tab:semantic_acc}), conducted across six datasets of varying complexity, shows that the full SEO-SLAM pipeline (RG+D+A) achieves the highest F1-scores in most cases and reduces false positive landmarks. 
Notably, in MD1, the RG+A pipeline outperforms SEO-SLAM, suggesting that in certain cases, a single-data association approach may be sufficient. In some runs (\textit{e.g.,} LG1 and LG2), certain objects (like a teacup or hammer) were never detected by the pipeline due to occlusions or low detection scores, resulting in fewer total landmarks.
The baseline RG configuration, which lacks both MLLM agents and multi-data association, consistently shows the lowest accuracy, demonstrating the efficacy of these components in improving the quality of semantic maps.

From the APE evaluation experiments~(Table~\ref{tab:tum}), while ORB-SLAM3~\cite{campos2021orb} maintains lower APE values overall, SEO-SLAM achieves a lower maximum APE in fr1desk2, indicating that its landmark refinement strategy helps mitigate outlier errors. However, its slightly higher mean APE values across all sequences are likely due to the initial false positives from object detections. Despite this, SEO-SLAM retains comparable localization accuracy, demonstrating that its emphasis on semantic richness does not significantly degrade trajectory estimation. Additionally, the qualitative evaluation results~(Fig.~\ref{fig:stata}) show that SEO-SLAM successfully tracks the overall structure of the trajectory. The floor plan overlay confirms that the estimated trajectory closely follows the expected path, validating the system’s capability to handle large-scale, multi-floor environments.


With an Nvidia 3090 GPU with an AMD Ryzen 7 5800X CPU, our SLAM pipeline runs at $\approx 6$ $fps$, with the object detector being the primary computational bottleneck. To assess the computational efficiency of our agents' scheme, we compare runtime performance across multiple configurations (Table~\ref{tab:time}) using the LG1 dataset over a 180-second period. Our scheme achieves a processing speed of 0.6207 sec/landmark, significantly outperforming a heterogeneous synchronous approach (1.5000 sec/landmark) and a synchronous GPT-4o setup (1.5517 sec/landmark). These results demonstrate that our asynchronous scheme with heterogeneous agents substantially enhances runtime efficiency. While the agents do not need to process every frame, further reducing inference time could improve their responsiveness and overall system performance.

\input{figs/api_demo}

\subsection{Applications}
One potential application of SEO-SLAM is in assistive robotics, where rich semantic mapping enables informed decision-making.
For example, as illustrated in Fig~\ref{fig:app}, a user wishing to go hiking can query an MLLM agent for footwear recommendations. The agent then 
consults SEO-SLAM’s up-to-date map (Fig~\ref{fig:intro}) to receive precise object details, including location and attributes, identify relevant objects, and provide targeted feedback.
This capability is particularly valuable for visually impaired users and can be extended to broader domains such as home assistance, where accurate semantic and spatial knowledge of objects is essential.

%% file: figs/results2_tab.tex
\begin{table}[t!]
\vspace{2mm}
\centering
\caption{Semantic Mapping Evaluation Results*\tnote{*}}
\label{tab:semantic_acc}
\begin{tabular}{ccccccc}
\toprule
\multirow{2}{*}{\textbf{Dataset}} & \multirow{2}{*}{\textbf{Method}}  & \multicolumn{3}{c}{\textbf{Semantic Acc.}} & \multicolumn{2}{c}{\textbf{Num. of Landmarks}} \\ 
\cmidrule(lr){3-5} \cmidrule(lr){6-7}
 &  & \textbf{P}$^\uparrow$ & \textbf{R}$^\uparrow$ & \textbf{F1}$^\uparrow$ & \textbf{Est.} & \textbf{False Pos.}$^\downarrow$ \\ 
\midrule

\multirow{3}{*}{SM1} & RG+D+A  & \textbf{0.83} & \textbf{1} & \textbf{0.91} & 12 & 2 \\
                     & RG+A & 0.89 & 0.80 & 0.84 & 9 & 1 \\
                     & RG+D & 0.69 & 0.90 & 0.78 & 13 & 4 \\
                     & RG & 0.56 & 0.50 & 0.53 & 9 & 4 \\
\midrule
\multirow{3}{*}{SM2} & RG+D+A  & \textbf{0.90} & \textbf{0.90} & \textbf{0.90} & 10 & 1 \\
                     & RG+A & \textbf{0.90} & \textbf{0.90} & \textbf{0.90} & 10 & 1 \\
                     & RG+D & 0.64 & \textbf{0.90} & 0.75 & 14 & 5 \\
                     & RG & 0.60 & 0.60 & 0.60 & 10 & 4 \\
\midrule
\multirow{3}{*}{MD1} & RG+D+A & 0.68 & 0.65 & 0.67 & 19 & 6 \\
 & RG+A & \textbf{0.83} & {0.75} & \textbf{0.79} & 18 & 3 \\
 & RG+D & 0.68 & 0.75 & 0.71 & 22 & 7 \\
 & RG & 0.61 & \textbf{0.85} & 0.71 & 28 & 11 \\
\midrule
\multirow{3}{*}{MD2} &  RG+D+A & \textbf{0.89} & 0.76 & \textbf{0.82} & 18 & 2 \\
 & RG+A & 0.69 & {0.86} & 0.77 & 26 & 8 \\
 & RG+D & 0.82 & 0.67 & 0.74 & 17 & 3 \\
 & RG & 0.70 & \textbf{0.90} & 0.79 & 27 & 8 \\
\midrule
\multirow{3}{*}{LG1} & RG+D+A & \textbf{0.86} & \textbf{0.61} & \textbf{0.72} & 22 & 3 \\
 & RG+A & 0.74 & 0.45 & 0.56 & 19 & 5 \\
 & RG+D & 0.80 & 0.52 & 0.63 & 20 & 4 \\
 & RG & 0.76 & 0.52 & 0.62 & 21 & 5 \\
\midrule
\multirow{3}{*}{LG2} & RG+D+A & \textbf{0.84} & \textbf{0.72} & \textbf{0.78} & 25 & 4 \\
& RG+A & 0.82 & 0.62 & 0.71 & 22 & 4 \\
 & RG+D & 0.62 & 0.52 & 0.57 & 24 & 9 \\
 & RG & 0.60 & 0.52 & 0.56 & 25 & 10 \\ 
\bottomrule
\end{tabular}
\begin{tablenotes}\footnotesize
\item[*] *P: Precision, R: Recall, F1: F-1 score, Est.: the number of estimated landmarks, False Pos.: the number of false positive landmarks.
\end{tablenotes}
\vspace{-4mm}
\end{table}

%% file: figs/results_tum.tex
\begin{table}[t!]
\vspace{2mm}
\centering
\caption{APE (m) results for the TUM RGBD Dataset}
\begin{tabular}{ccccccc}
\toprule
Dataset & Method  &  Max$^\downarrow$& Mean$^\downarrow$ & Median$^\downarrow$ & RMSE$^\downarrow$ \\ 
\midrule
\multirow{2}{*}{fr1desk} & ORB-SLAM3 & 0.0624 & 0.0139 & 0.0112 & 0.0169 \\ 
                     & Ours  & 0.0544  & 0.0182 & 0.0147 & 0.0217 \\ 
\midrule
\multirow{2}{*}{fr1desk2} & ORB-SLAM3 & 0.0723 & 0.0245 & 0.0221 & 0.0284 \\ 
                     & Ours  & 0.0477  & 0.0216 & 0.0201 & 0.0237 \\ 
\midrule
\multirow{2}{*}{fr2desk} & ORB-SLAM3 & 0.0375 & 0.0165 & 0.0149 & 0.0180 \\ 
                     & Ours  & 0.0526 & 0.0267 & 0.0234 & 0.0299 \\ 
\bottomrule
\end{tabular}
\label{tab:tum}
\vspace{-4mm}
\end{table}

%% file: figs/stata_qual.tex
\begin{figure}[t!]
\vspace{2mm}
        \centering\includegraphics[width=1.0\linewidth]{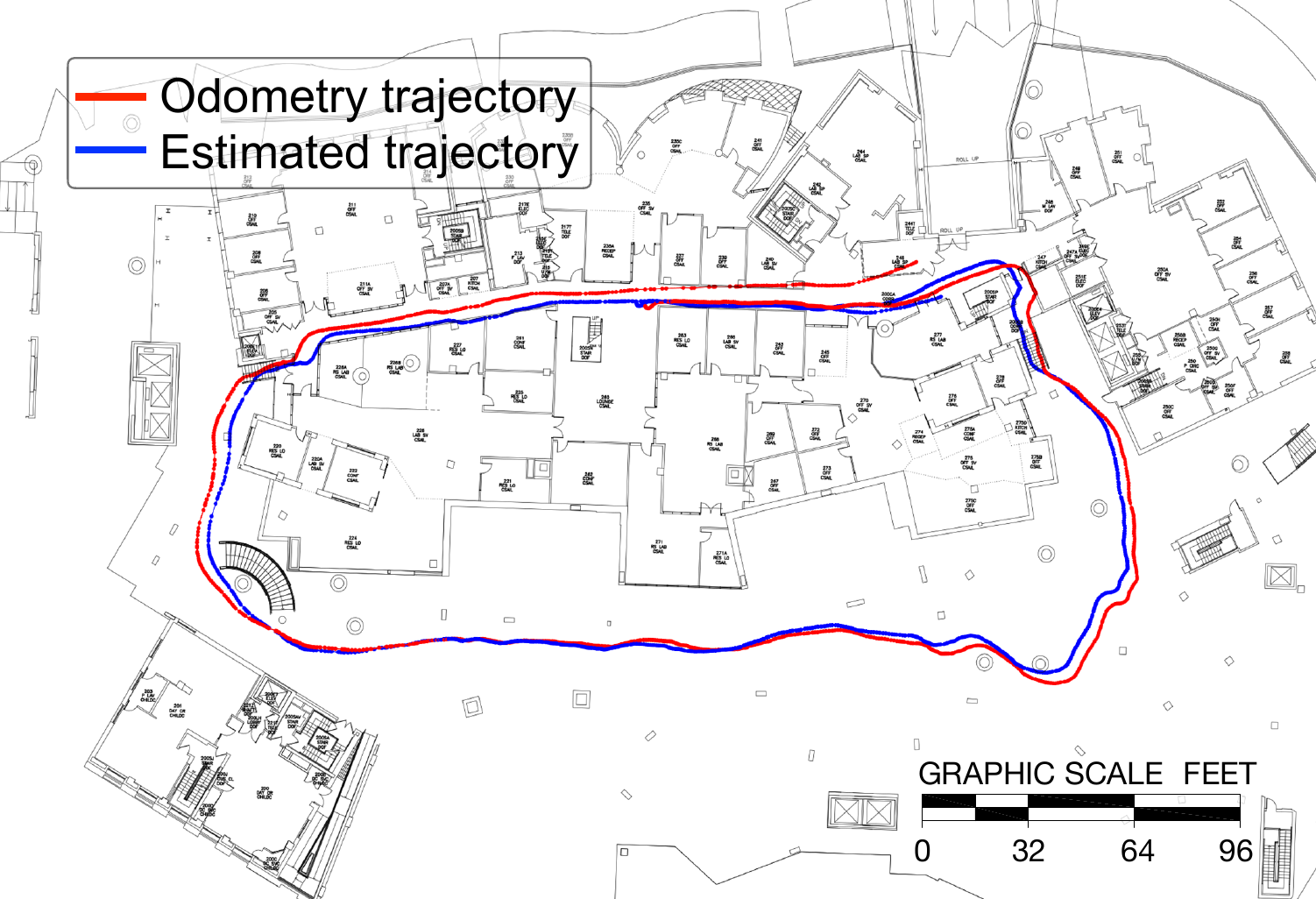}
        \label{fig:image2}
        \vspace{-5mm}
    \caption{Qualitative evaluation of SEO-SLAM on a trajectory spanning the 1st and 2nd floors of the MIT Stata Center. Both estimated (blue) and odometry (red) trajectories are overlaid on a floor plan.}
    \label{fig:stata}
    \vspace{-6mm}
\end{figure}

%% file: figs/results_time.tex
\begin{table}[t!]
\vspace{2mm}
\centering
\caption{Comparison of Landmark Processing Time Across Different Configurations}
\begin{tabular}{%
  >{\centering\arraybackslash}m{2.2cm} 
  >{\centering\arraybackslash}m{2.0cm} 
  >{\centering\arraybackslash}m{2.5cm} 
}
\toprule
 & No. of processed landmarks$^\uparrow$ & Avg. processing time (sec/landmark)$^\downarrow$ \\ 
\midrule
\textbf{Async, Llama, GPT-4o (ours)} & 290 & 0.6207 \\
\midrule
Sync, Llama, GPT-4o & 120 & 1.5000 \\
\midrule
Sync, GPT-4o & 116 & 1.5517 \\
\bottomrule
\end{tabular}
\label{tab:time}
\vspace{-4mm}
\end{table}

%% file: figs/api_demo.tex
\begin{figure}[t!]
\vspace{2mm}
        \centering
        \includegraphics[width=0.9\linewidth]{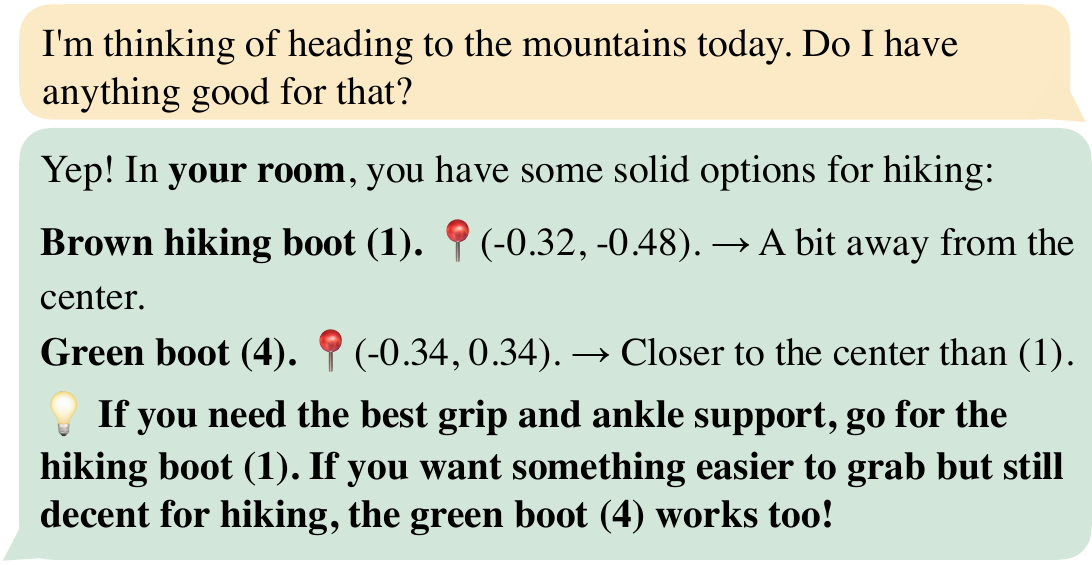}
        \caption{A user queries GPT-4o regarding suitable hiking footwear, using SEO-SLAM’s semantic map from Fig~\ref{fig:intro}. GPT-4o offers context-aware recommendations with spatial locations.
        }
        \vspace{-4mm}
        \label{fig:app}
\end{figure}

%% file: src/7-conclusion.tex
\section{Conclusions}

We present SEO-SLAM, a novel object-level semantic SLAM framework that enhances mapping accuracy by integrating multimodal large language model (MLLM) agents. 
By leveraging open-vocabulary object detection and heterogeneous MLLM agents, our system refines landmark labels, mitigates perceptual aliasing, and adapts to scene changes. The proposed asynchronous MLLM agent framework ensures efficient label verification, description generation, and landmark merging without compromising SLAM runtime. Experimental results demonstrate that SEO-SLAM significantly improves semantic accuracy, reduces false positive landmarks, and maintains robust data association in cluttered and scene-changing environments. Furthermore, our system achieves comparable localization accuracy while producing semantically rich maps. This can benefit downstream tasks such as robotic assistance with object-centric scene understanding, as demonstrated in our study.
